\documentclass[letterpaper, 10 pt, journal, twoside]{IEEEtran}  

\usepackage{graphicx} 
\graphicspath{{figures/}}

\usepackage{amsthm}
\usepackage{amsmath} 
\usepackage{amssymb}  
\usepackage{mathtools}
\usepackage{ulem} 
\usepackage{multirow}
\usepackage{verbatim}
\usepackage{soul}
\usepackage{algorithm}
\usepackage{algpseudocode}

\usepackage{color}
\usepackage{latexsym}
\usepackage{multicol}

\usepackage{enumitem}
\usepackage[font=small,labelfont=bf]{caption}
\usepackage{lipsum}
\usepackage{subcaption}



\renewcommand{\thefigure}{\arabic{figure} }

\IEEEoverridecommandlockouts                              

\begin{document}

\title{\LARGE \bf
Multi-Abstractive Neural Controller: An Efficient Hierarchical Control Architecture for Interactive Driving
}

\author{Xiao Li$^1$, Igor Gilitschenski$^4$, Guy Rosman$^3$, Sertac Karaman$^2$ and Daniela Rus$^1$


\thanks{$^{1}$ Computer Science and Artificial Intelligence Lab, Massachusetts Institute of Technology
         {\tt\small \{xiaoli, rus\}@mit.edu}}%
\thanks{$^{2}$ Laboratory for Information and Decision Systems, Massachusetts Institute of Technology
        {\tt\small sertac@mit.edu}}%
\thanks{$^{3}$ Toyota Research Institute
        {\tt\small \{guy.rosman\}@tri.global}}%
\thanks{$^4$ University of Toronto \{\tt\small gilitschenski\}@cs.toronto.edu }
}

{



\maketitle

\begin{abstract}
As learning-based methods make their way from perception systems to planning/control stacks, robot control systems have started to enjoy the benefits that data-driven methods provide. Because control systems directly affect the motion of the robot, data-driven methods, especially black box approaches, need to be used with caution considering  aspects such as stability and interpretability. In this paper, we describe a differentiable and hierarchical control architecture. The proposed representation, called \textit{multi-abstractive neural controller}, uses the input image to control the transitions within a novel discrete behavior planner (referred to as the visual automaton generative network, or \textit{vAGN}). The output of a vAGN controls the parameters of a set of dynamic movement primitives which provides the system controls. We train this neural controller with real-world driving data via behavior cloning and show improved explainability, sample efficiency, and similarity to human driving.
\end{abstract}
\section{INTRODUCTION}

With robotic and autonomous driving applications expanding from structured environments (factory floors, warehouses, etc) to open environments (road, homes, etc), traditional optimal planning/control methods will be insufficient in handling the large variety of edge-cases as manual specifications for them is intractable. Data-driven methods such as imitation learning have shown promising results in learning generalizable and capable robot control policies from human demonstrations. However, enabling blackbox policies such as neural networks to consistently produce stable behaviors outside of the training data distribution remains a challenge, hindering their adoption in safety-critical applications such as autonomous driving.  In this work, our aim is to address the following question: ``\textit{can we design a robot control policy representation that (1) is explainable in its decision making process, (2) is resilient to unstable behaviors , and (3) is trainable end-to-end using expert demonstrations?}"

Combining model-based planning and control with learning components allows the robot system to have the stability and safety properties of model-based controllers while complexity and uncertainty of the environment (elements that are challenging to manually and exhaustively integrate into model/rule-based components) can be delegated to the data-driven components. Most efforts in this space integrate learning into a certain component of a motion planner including the cost function \cite{Zeng2019EndToEndIN}, dynamics \cite{Salzmann2022NeuralMPCDL}, the solver \cite{learnmppi,Ichnowski2021AcceleratingQO} and constraints \cite{Dawson2022SafeCW}. Less work has been done to integrate learning one level up the planning stack - into the behavior planner. The concept of a multi-abstractive neural controller provides an initial effort to fill this gap and expand the learnability of a robotic planning/control stack into the high-level discrete planning domains.

In this paper, we start by referring to the classical planning/control stack where a discrete behavior planner (commonly in the form of a finite state machine) feeds high-level decisions (along with additional environment feedback) into a motion planner, which in turn outputs controls for the dynamic system. \textbf{Our approach is to design a differentiable architecture of this stack, and define their interfaces such that its structure can be learned from data}. We refer to this representation as the \textit{multi-abstractive neural controller}. Figure \ref{fig:intro_description} illustrates the desired architecture.

For the behavior planner, we introduce the visual automaton generative network (vAGN) - a differentiable automaton that takes in visual features and learns its transition structure from demonstration data. We adopt dynamic movement primitives (DMP) \cite{Ijspeert2002MovementIW} as the motion controller (for its stability properties and simplicity) and introduce a novel method that interfaces vAGN with DMP that significantly improves the stability of the controlled system. To summarize our contributions, we 
\begin{itemize}
    \item introduce vAGN - an automaton planner with learnable latent structure;
    \item introduce the multi-abstractive neural controller - a hierarchical robot control representation that interfaces vAGN with DMP through DMP parameter control;
    \item demonstrate in a real world driving dataset that the proposed neural robot controller achieves high sample efficiency while balancing safety, optimality and comfort.
\end{itemize}

\begin{figure*}[htb]
    \centering
    \includegraphics[width=2.\columnwidth]{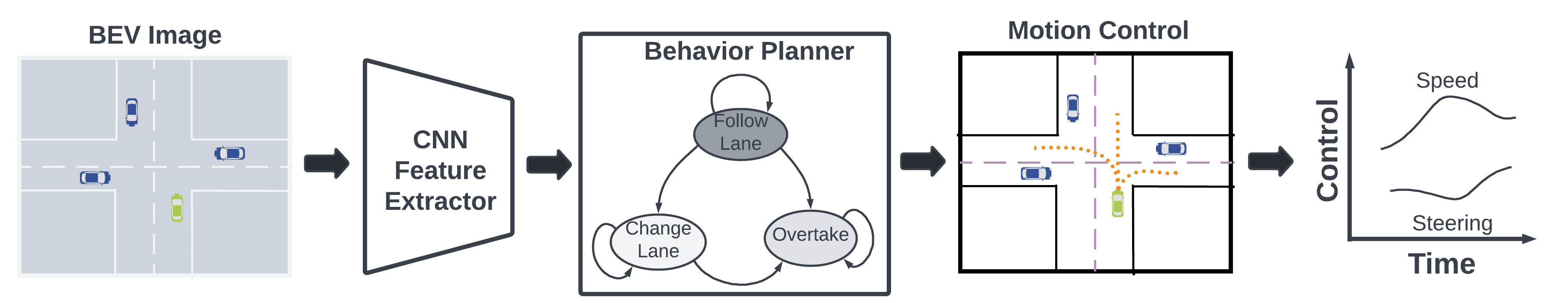}
    \caption{\textbf{A differentiable planning and control representation.} By designing the discrete behavior planner and motion controller as differentiable components and connecting them such that their structure can be learned from demonstrations, both generalizability from data-driven methods and stability from model-based methods can be achieved with a more efficient control representation.}
    \label{fig:intro_description}
\end{figure*}

\noindent In this work, we focus on applying the neural controller on autonomous driving applications. 

\section{RELATED WORK}\label{sec:literature review}
\textbf{Joint learning and planning.} Constructing learnable planners have been looked at in the past. The authors of \cite{Meng2019NeuralAN} use a ResNet50 to extract features from camera images which are used to predict the acceleration and metric components of Riemannian motion policy (RMP). Similarly, \cite{Kabzan2019LearningBasedMP} use learning components to generate parameters the model predictive controller. In \cite{Zeng2019EndToEndIN}, the authors use a neural network to generate cost maps from LiDAR and map inputs which are used to rank trajectory samples. In \cite{Salzmann2022NeuralMPCDL}, the authors use neural networks to approximate the dynamics used in an model predictive controller (MPC). On a different line of idea, the authors of \cite{learnmppi} propose to learn the update rules for the model predictive path integral (MPPI) planner. In \cite{prm-rl}, the authors use reinforcement learning to learn short horizon policies and probabilistic road map for global navigation. The authors of \cite{Fox2019Multi} learn hierarchical task planners using imitation learning, but their planner structures are either untrainable or hard to interpret.  Most of the work in the this space aim to learn certain component of a motion planner (dynamics, cost, optimizer, etc), but little work is done to develop learnable (discrete) behavior planners. Our work fills this gap by introducing a differentiable automaton with learnable transition structure (often a challenge to manually design). Then using CNN as the input processing unit, it is common practice to visualize its feature maps for explainability purposes. The class activation map \cite{class_activation_map,Bojarski2016VisualBackPropVC} is one such method which we have adopted in this work. In addition, the visual backpropagation method \cite{Bojarski2016VisualBackPropVC, Lechner2020NeuralCP} is also commonly used in self-driving applications.   

\textbf{Neural state machines.} While little work has been done in learnable behavior planners, differentiable state machines have been developed in the field of visual question answering (VQA) and natural language processing (NLP). In \cite{Hudson2019LearningBA, Kochiev2021NeuralSM}, the authors propose the neural state machine - a scene graph constructed from images that is able to make inferences based on natural language instructions. In \cite{Hannun2020DifferentiableWF}, the authors introduce the differentiable weighted finite-state transducers to express and design sequence-level loss functions (used in handwriting and speech recognition). The authors of \cite{Ke2022LearningTI} proposes an approach to learn causal Bayesian networks from data. And finally, the authors of \cite{Starke2019NeuralSM} learn a neural state machine used for character-scene interactions. Because these works are not tailored to robotic planning applications, they lack one or more of the following features that prevents them from being readily available as behavior planners, (a) the inability to learn state transition structures (b) the requirement of having ground truth graphs (or data-to-graph distribution) as supervision, (c) not using an underlying graph structure (\cite{Starke2019NeuralSM} uses a 3 layer fully connected gating network).  Our work addresses these problem and is demonstrated to be effective in the self-driving domain. The neural hybrid automaton (NHA) proposed by \cite{Poli2021NeuralHA} learns a hybrid control system and is perhaps the closest to our work. However, NHA requires state estimates (e.g. position, speed, etc) as input whereas our method takes images as inputs.
Explainability for general machine learning methods is discussed in \cite{Lipton2016TheMO,Narayanan2018HowDH}. In this work, we focus on explainable learning systems for planning and control.

\section{BACKGROUND}
\subsection{Linear Dynamic Movement Primitives (DMP)}
A DMP \cite{Ijspeert2002MovementIW} consists of a second order point attractor system added with a forcing function as below
\begin{equation}
\label{eq:dmp}
\tau \ddot{\boldsymbol y}  =\alpha_{y}\left(\beta_{y}(\boldsymbol g- \boldsymbol y)-\dot{\boldsymbol y}\right)+\boldsymbol f(s,x|\theta_y), \quad \tau \dot{x} = \alpha_x x
\end{equation}

\noindent where $g$ is the goal state; $\alpha_{y}$ and $\beta_{y}$ define the behavior of the second order system;$\tau$ is a time constant; $x$ is a phase variable controlling the influence of the forcing function on the point attractor system. Appropriately setting $\tau, \alpha_{y}, \beta_{y}$, the convergence of the underlying dynamic system to $y=g$ is ensured \cite{Ijspeert2013DynamicalMP} and the system is stable with respective to the goal. The first part of Equation \eqref{eq:dmp} is often referred to as the transformation system. The transformation system serves to stably guide the robot to the goal with a trajectory jointly controlled by the point attractor and the forcing function. The second part (first order system of $x$) is the canonical system which controls the decay of x and hence reduces the effect of the forcing function as the robot gets close to the goal. $f(s,x|\theta_y)$ ($s$ can be additional state information) is a learnable forcing function often in the form of a linear combination of $N$ nonlinear Radial Basis Functions (RBFs). This allows the robot to reach the goal state by following a desired path influenced by $f(\cdot)$.

\subsection{Quaternion Dynamic Movement Primitives}
The DMP introduced in the previous section generates only linear movement. As orientation is just as important in defining robot motion, the authors of \cite{AbuDakka2015AdaptationOM} introduced the equivalent of Equations (1) and (2) for unit quaternions $\mathtt{q}=[v,\mathbf{u}] \in$ $\mathcal{S}^{3}$ ($\mathcal{S}^{3}$ is a unit sphere in $\mathbb{R}^{4}, v \in \mathbb{R}$, and $\mathbf{u} \in \mathbb{R}^{3}$). as follows

\begin{equation}
\label{eq:qdmp}
\begin{split}
\tau \dot{\boldsymbol{\eta}} &=\alpha_{q}\left(\beta_{q} 2 \log ^{q}\left(\boldsymbol{g}_{q} * \overline{\mathtt{q}}\right)-\boldsymbol{\eta}\right)+\mathbf{f}_{q}(s,x|\theta_q) \\
\quad \tau \dot{\mathtt{q}} &=\frac{1}{2} \boldsymbol{\eta} * \mathtt{q}
\end{split}
\end{equation}

\noindent where $\mathbf{g}_{q} \in \mathcal{S}^{3}$ denotes the goal orientation, the quaternion conjugation is defined as $\overline{\mathtt{q}}=\overline{v+\mathbf{u}}=v-\mathbf{u}$, and $*$ denotes the the quaternion product

\begin{equation}
\begin{split}
\mathtt{q}_{1} * \mathtt{q}_{2} &=\left(v_{1}+\mathbf{u}_{1}\right) *\left(v_{2}+\mathbf{u}_{2}\right)\\
&=\left(v_{1} v_{2}-\mathbf{u}^{\top}{ }_{1} \mathbf{u}_{2}\right)+\left(v_{1} \mathbf{u}_{2}+v_{2} \mathbf{u}_{1}+\mathbf{u}_{1} \times \mathbf{u}_{2}\right)
\end{split}
\end{equation}

\noindent $\boldsymbol{\eta} \in \mathbb{R}^{3}$ is the scaled angular velocity $\boldsymbol{\omega}$ and treated as unit quaternion with zero scalar $(v=0)$. The function $\log ^{q}(\cdot): \mathcal{S}^{3} \mapsto \mathbb{R}^{3}$ is given as

\begin{equation}\label{eq:log_quat}
\log ^{q}(\mathtt{q})= \begin{cases}\arccos (v) \frac{\mathbf{u}}{\|\mathbf{u}\|}, & \mathbf{u} \neq \mathbf{0} \\
{\left[\begin{array}{lll}
0 & 0 & 0
\end{array}\right]^{\top},} & \text { otherwise }\end{cases}
\end{equation}

\noindent where $\|\cdot\|$ denotes $\ell_{2}$ norm. Equation (\ref{eq:log_quat}) can be integrated as
\begin{equation}
\mathtt{q}(t+\delta t)=\operatorname{Exp}^{q}\left(\frac{\delta t}{2} \frac{\boldsymbol{\eta}(t)}{\tau}\right) * \mathtt{q}(t),
\end{equation}

\noindent where $\delta_{t}>0$ denotes a small constant. The function $\operatorname{Exp}^{q}(\cdot): \mathbb{R}^{3} \mapsto \mathcal{S}^{3}$ is given
\begin{equation}
\operatorname{Exp}^{q}(\boldsymbol{\omega})= \begin{cases}\cos (\|\boldsymbol{\omega}\|)+\sin (\|\boldsymbol{\omega}\|) \frac{\boldsymbol{\omega}}{\|\boldsymbol{\omega}\|}, & \boldsymbol{\omega} \neq \mathbf{0} \\
1+\left[\begin{array}{lll}
0 & 0 & 0
\end{array}\right]^{\top}, & \text { otherwise. }\end{cases}
\end{equation}

\noindent Both mappings become one-to-one, continuously differentiable and inverse to each other if the input domain of the mapping $\log ^{q}(\cdot)$ is restricted to $\mathcal{S}^{3}$ except for $-1+\left[\begin{array}{lll}0 & 0 & 0\end{array}\right]^{\top}$, while the input domain of the mapping $\operatorname{Exp}^{q}(\boldsymbol{\omega})$ should fulfill the constraint $\|\boldsymbol{\omega}\|<\pi$.

\section{Multi-Abstractive Neural Controller}

In this section, we introduce the general architecture of the multi-abstractive neural controller with its three main component, (1) a CNN-based visual predicate extractor that learns and outputs visual features meaningful for high-level behavior planning; (2) a differentiable graphical planner with learnable structure and (3) a DMP with parameters controlled by the output of the behavior planner. The overall architecture is illustrated in Figure \ref{fig:architecture}. we refer to our controlled vehicle as the \textit{ego} vehicle, and vehicles in traffic as the \textit{ado} vehicles.

\begin{figure*}[htb]
    \centering
    \includegraphics[width=2\columnwidth]{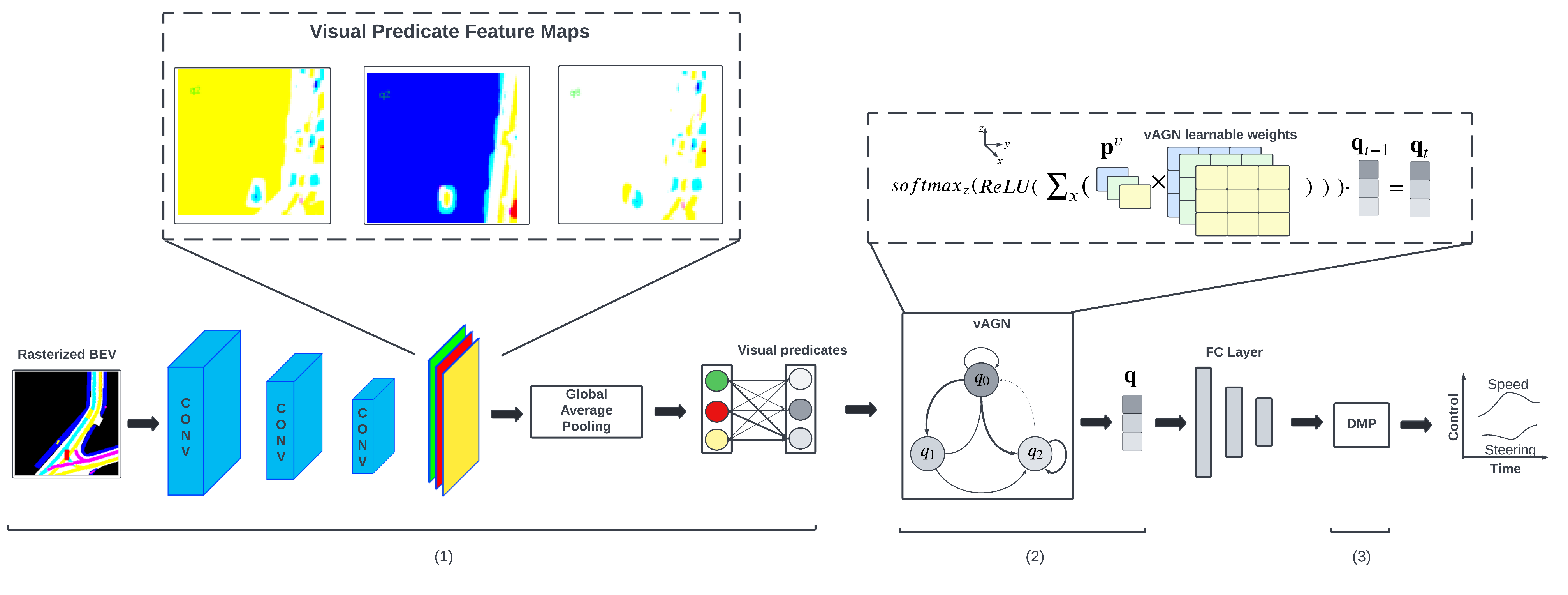}
    \caption{\textbf{Multi-abstractive neural controller.} The architecture contains three components,  (1) a CNN-based visual predicate extractor that learns and outputs visual features meaningful for high-level behavior planning; (2) a differentiable graph-based planner (vAGN) with learnable structure and (3) a DMP with parameters controlled by the output of vAGN.}
    \label{fig:architecture}
\end{figure*}

\subsection{Visual Predicate Extractor}
\label{sec:vp extractor}

The purpose of the visual predicate extractor is to learn a feature vector where each of its element corresponds to the existence of a semantic feature in the input image (e.g. lane, stop sign, etc). Let $\mathcal{X}$ be the input image, here we use a bird's eye view image (BEV) with traffic components semantically colored. We perform the following operations to obtain the visual predicate feature vector
\begin{equation}\label{eq:visual predicate extractor}
\begin{split}
    \boldsymbol f^v &= \texttt{ConvEncoder}(\mathcal{X})\\
    \boldsymbol p^v &= \texttt{Linear}\big(\texttt{GlobalAveragePooling}(\boldsymbol f^v)\big).
\end{split}
\end{equation}

In the equation above, $\texttt{ConvEncoder}(\cdot)$ is a set of convolution layers applied to the image. Its output is a set of feature maps to which we apply global average pooling \cite{Lin2014NetworkIN} to. This serves to identify whether certain components exists in the image (i.e. pedestrians, intersections, etc). Lastly we apply a linear transformation to the output of the pooling layer. This allows each element in $\boldsymbol p^v$ to contain a weighted sum of all the identified features, which provides richer information to the downstream planner. The architecture of this component is inspired by the class activation map \cite{Zhou2016LearningDF}. As shown in Figures 2 and 3, the visual predicates constitute the explainability components of vAGN that serves to illustrate the neural controller's internal decision making process.

\subsection{Visual Automaton Generative Network (vAGN)}
\label{sec:vAGN}

Given the visual predicates $\boldsymbol p^v \in {\rm I\!R}^M$ and the number of automaton nodes $N$ (as a hyperparameter), the current automaton state $\boldsymbol q_t \in {\rm I\!R}^N$ is represented
as an $N$-vector with each entry corresponding to the probability of being in $q_i$ ($\boldsymbol q$ can be seen as the hidden state equivalent of an LSTM). As an example, for an automaton with 3 states $[q_1,q_2,q_3]$, $\boldsymbol q=[0.4,0.3,0.3]$ is the state distribution of this probabilistic graph. The learnable parameters of vAGN is its set of weighted transition matrices $\mathcal{W} \in {\rm I\!R}^{M\times N \times N}$. The vAGN update law is as follows
\begin{equation}
\label{eq:vagn wpv}
    \mathcal{W}^{p^{v}} = \sum_{i \in \{0, M-1\}} (\boldsymbol p^v_i \times W_i)
\end{equation}
\begin{equation}\label{eq:vagn}
    \boldsymbol q_{t} = \underset{\textrm{along column}}{\textrm{softmax}}(\textrm{ReLU}(\mathcal{W}^{p^{v}})) \cdot \boldsymbol q_{t-1}
\end{equation}

The dimension of $\mathcal{W}$ is such that each element of $p^v_i \in \boldsymbol p^v$ has a corresponding $W_i \in \mathcal{W} \in {\rm I\!R}^{N\times N}$. In Equation \eqref{eq:vagn wpv},  $\mathcal{W}^{p^v} \in {\rm I\!R}^{N\times N}$ indicates the combined influence of the visual predicates on the transition of $\boldsymbol q$ probabilities. In Equation \eqref{eq:vagn}, we first apply a $\textrm{ReLU}(\cdot)$ to $\mathcal{W}^{p^v}$ to preserve only the positive transitions. Softmax is applied to ensure that the columns of $\mathcal{W}^{p^v}$ sum to one (the sum of probabilities of transitioning out of any $q$ state to another state is 1).  Finally, the resultant transition matrix is applied to $\boldsymbol q_{t-1}$ via dot product. 

\subsection{Interfacing vAGN with DMP}
\label{sec:vAGN_DMP}

The most common way of learning with DMPs is to learn the parameters of the forcing functions $\boldsymbol f(\cdot)$ and $\boldsymbol f_q(\cdot)$ in Equation \eqref{eq:dmp}. This works well with finite horizon tasks where the effects of the forcing functions diminish over time (with the canonical system) and the system eventually reaches the goal state. However, for potentially long/infinite horizon tasks (such as driving), it is less obvious how the learned forcing functions should scale. If they are always kept in effect, then the system may never reach the goal state. As an alternative, instead of adding the forcing function with point attractor system, we propose to use the forcing function to learn parameters of the point attractor system. As a result, Equations \eqref{eq:dmp} becomes
\begin{equation}
\label{eq:dmp2}
\begin{split}
\ddot{\boldsymbol y}  &= \alpha_{y}(s|\theta_y)\left(\beta_{y}(s|\theta_y)(\boldsymbol g- \boldsymbol y)-\dot{\boldsymbol y}\right) \\
\dot{\boldsymbol{\eta}} &=\alpha_{q}(s|\theta_q)\left(\beta_{q}(s|\theta_q) 2 \log ^{q}\left(\boldsymbol g_{q} * \overline{\mathtt{q}}\right)-\boldsymbol{\eta}\right),
\end{split}
\end{equation}

\noindent where $\{\alpha_{y}(s|\theta_y), \beta_{y}(s|\theta_y), \alpha_{y}(s|\theta_q), \beta_{y}(s|\theta_q)\}$ are the learned functions (the time constant $\tau$ can be incorporated in these 4 parameters). This can then be easily connected with the output of vAGN by 
\begin{equation}\label{eq:dmp_final}
    \begin{bmatrix}
    \ddot{\boldsymbol y} \\
    \dot{\boldsymbol \eta} 
    \end{bmatrix} = \texttt{DMP}(\texttt{FC}(\boldsymbol q), \boldsymbol g, \boldsymbol g_q)
\end{equation}

\noindent where $\texttt{FC}(\boldsymbol q) = \{\alpha_{y}, \beta_{y}, \alpha_{y}, 
\beta_{y}\}$ is a set of fully connected layers transforming the vAGN states 
to the DMP parameters. The parameters $\theta_y, \theta_q$ becomes the 
upstream network parameters (those of vAGN and CNN). Note that Equation 
\eqref{eq:dmp_final} outputs accelerations. We use linear and angular 
velocities as controls, therefore we numerically integrate Equation 
\eqref{eq:dmp_final} once to obtain the final output. Having the output 
of vAGN control the parameters of the DMP point attractor system run the risk 
of rendering the tracking behavior unstable. The point attractor system of the DMP can be rewritten as 
    
\begin{equation}\label{eq: second order system}
    \ddot{y} +\alpha \dot{y} +\alpha \beta y=0
\end{equation}

\noindent where the constant term is neglected (does not effect stability analysis). Equation \eqref{eq: second order system} is a typical second order system where stability is determined by

\begin{equation}\label{eq:zeta}
    \zeta =\frac{\alpha }{2\sqrt{\alpha \beta }}.
\end{equation}

\noindent The system can become unstable when $\displaystyle \zeta < 0$ and periodic when $\displaystyle \zeta \approx 0$. In our experiments, we have not run into this particular problem because we used the sigmoid activation when outputing $\displaystyle \alpha $ and $\displaystyle \beta $, $\displaystyle \zeta $ is always greater than zero. Fortunately, during training and deployment we have not run into the case where $\displaystyle \zeta \approx 0$. For the driving task that we are targeting, the ego vehicle is tracking a moving target along the center lane, and the parameters $\displaystyle \alpha ,\beta $ are time varying functions of the upstream network, therefore, the system under our neural controller is resilient to unstable $\displaystyle \alpha ,\beta $ values as long as they don't stay unstable for an extended period of time. To avoid $\displaystyle \zeta \approx 0$, one can set $\displaystyle \beta =\alpha /4$ such that the system is always critically damped (at the cost of somewhat limiting vehicle behaviors). One could also use an auxiliary loss at training time to prevent the upstream vAGN+FC from outputing $\displaystyle \alpha $ and $\displaystyle \beta $ that are too close together.

vAGN will thus serve to control the tracking ``aggressiveness".  We show later in the experiments that various driving maneuvers can 
emerge from this combination and will also discuss its limitations. Note also that \cite{Koutras2019ACF} introduces an alternative formulation of the forcing function that provides better stability. This does not affect our neural controller as we directly alter the point attractor's parameters ($\alpha, \beta$) and neglect the forcing function.  This is to avoid instabilities as a result of the forcing terms overturning the point attractor terms. Such overturning can lead to crashes in driving tasks (may cause less of a safety concern in manipulation tasks).

Algorithm \ref{alg:vagn} describes the procedures of learning a multi-abstractive neural
controller from demonstrations.

\begin{algorithm}
\caption{Learning with Multi-Abstractive Neural Controller}
\label{alg:vagn}
\begin{algorithmic}[1]
\State \textbf{Inputs}: number of vAGN nodes $N$; dataset $\boldsymbol X$; number of iterations $I$; learning rate $\gamma$
\State $\theta^{vAGN}, \theta^{CNN} \leftarrow \texttt{Initialize}(N)$ 
\For{i=1 \ldots I}
    \State Sample a minibatch of $m$ data samples $(\mathcal{X}, \boldsymbol g, \boldsymbol g_q, \boldsymbol a)$
    \State $\boldsymbol p^v = \texttt{VisualPredicateExtractor}(\mathcal{X})$ \Comment Equation \eqref{eq:visual predicate extractor}
    \State $\boldsymbol q^\prime = \texttt{UniformInit()} \textrm{ or } \texttt{RandomInit()}$
    \State $\boldsymbol q = \texttt{vAGN}(\boldsymbol p^v, \boldsymbol q^\prime)$ \Comment{Equation \eqref{eq:vagn}}
    \State $\hat{\boldsymbol a} = \texttt{DMP}(\texttt{FC}(\boldsymbol q), \boldsymbol g, \boldsymbol g_q)$ \Comment{Equation \eqref{eq:dmp_final}}
    \State $L = \texttt{MSE}(\hat{\boldsymbol a}, \boldsymbol a)$
    \State $(\theta^{vAGN},\theta^{CNN})\leftarrow (\theta^{vAGN}, \theta^{CNN}) - \gamma \frac{1}{m} \nabla L$
\EndFor
\end{algorithmic}
\end{algorithm}

\noindent In Algorithm 1, the loop $\textrm{for } i=1...I$ refers to iteration over minibatches of data (not over time in a rollout). During training, we do not know the groundtruth $\boldsymbol q$, therefore, we assume a uniform or random $\boldsymbol q$ to pass into the vAGN() module. This is a preliminary resolution and yields reasonable results, but can definitely be improved (for example by consuming groundtruth $\boldsymbol q$ from a high-level behavior filter)

\section{Experiments and Results}
\label{sec:experiments and results}

\begin{figure*}[htb]
    \centering
    \includegraphics[width=2.\columnwidth]{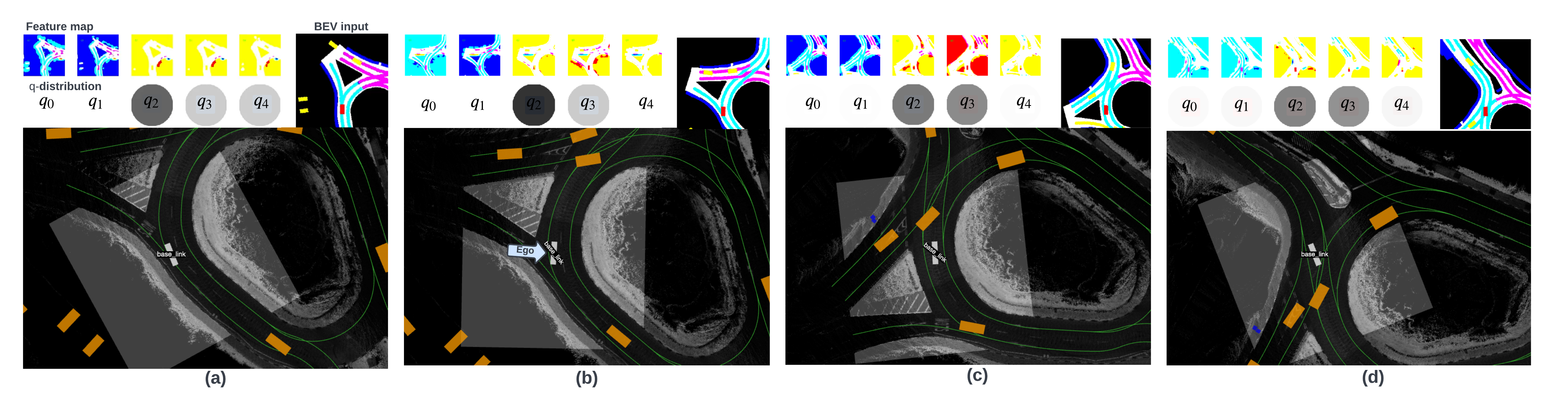}
    \caption{\textbf{An example execution trace.} Within each sub-figure we also show the current vAGN state distribution (darker the color means higher probability) and the saliency map indicating where each q-state is attending to. Color indicates the level of attention and ranks from high to low as: $\textrm{yellow} \rightarrow \textrm{red} \rightarrow \textrm{cyan} \rightarrow \textrm{magenta}$. To the right of each q-distribution plot we also show the semantically colored BEV image sent as input to the network.\textbf{(a)} Ego vehicle drives in an open area with no nearby ado vehicles, vAGN attends mainly to the road boundaries. \textbf{(b)} vAGN starts to notice more of ado vehicles. \textbf{(c)} The ado vehicles cut in front of the ego vehicle and vAGN shifts part of its focus to these vehicles. \textbf{(d)} The ego vehicle proceeds out of the roundabout and vAGN attends back to mainly road boundaries.}
    \label{fig:Execution traces}
\end{figure*}

\begin{figure*}
    \centering
    \includegraphics[width=1.3\columnwidth]{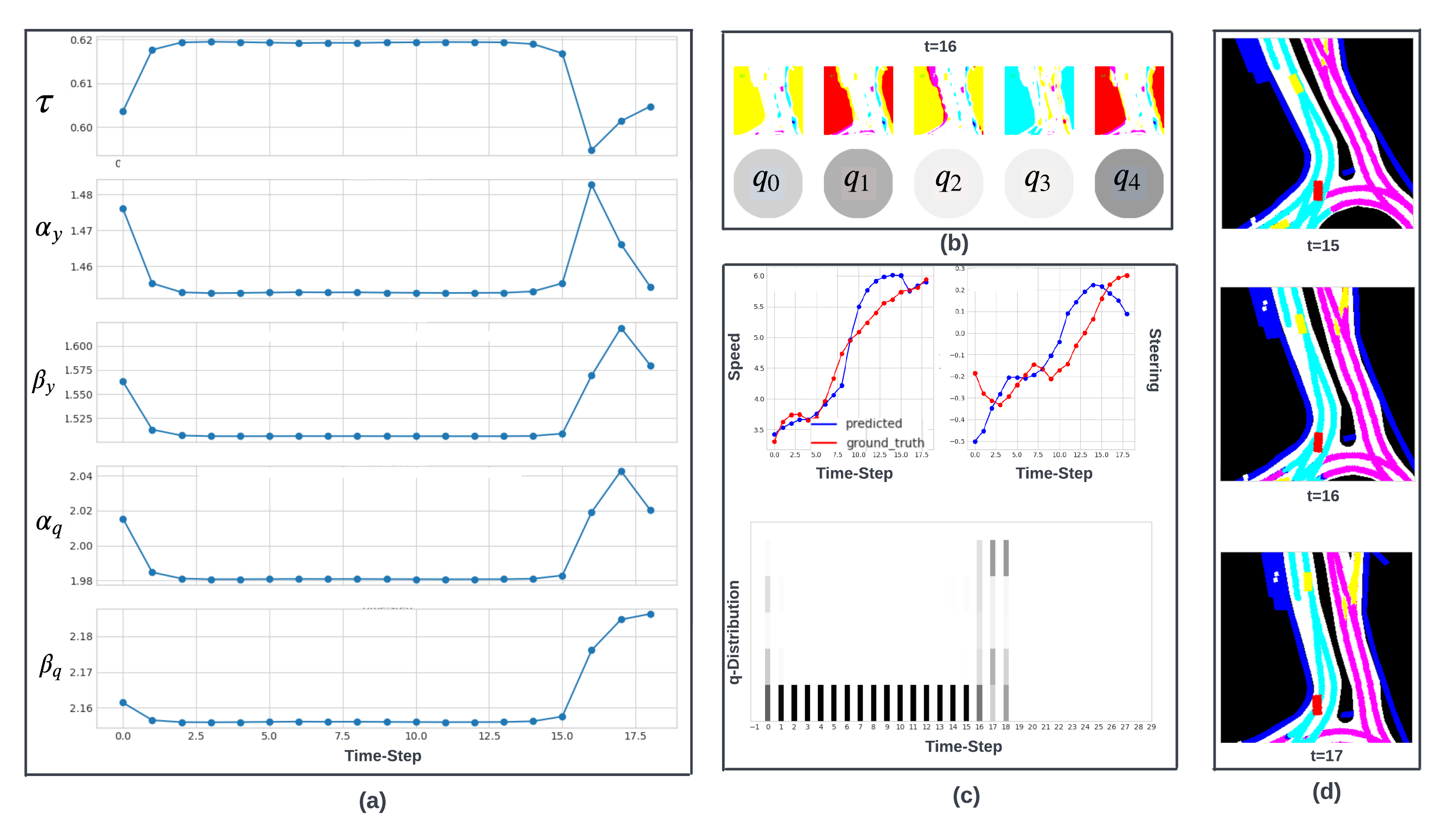}
    \caption{\textbf{Explainability traces.} \textbf{(a)} The evolution of DMP parameters over time. \textbf{(b)} The vAGN state distribution and corresponding saliency maps. \textbf{(c)} Control output from our neural controller (blue) and human controls (red) with synchronized q-distribution. \textbf{(d)} BEV image input at 3 timesteps. After step 15, the ego vehicle drives too close to the road boundary. As a result, vAGN shifts attention and the q-distribution to reduce the DMP parameters, which in turn reduces the speed and steering to prevent collision.}
    \label{fig:Explainability traces}
\end{figure*}

\begin{figure*}
    \centering
    \includegraphics[width=2.\columnwidth]{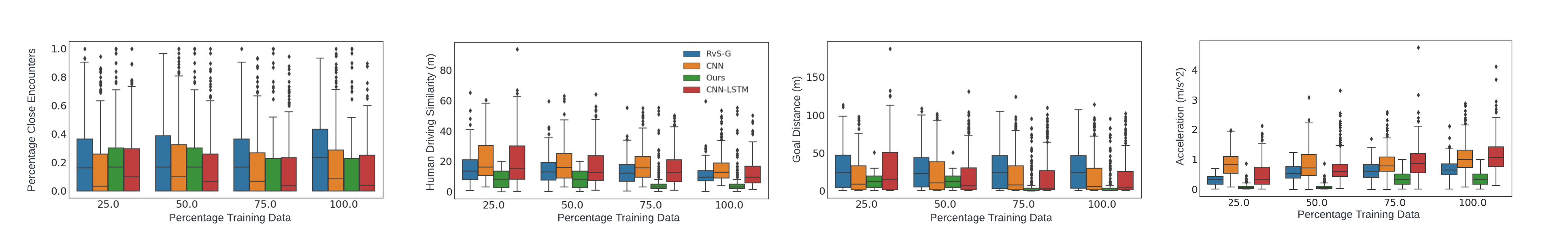}
    \caption{\textbf{Sample and model efficiency study.} Model performance trained at 25\%, 50\%, 75\% and 100\% training data. Our approach is able to achieve relatively high human driving similarity (low ADE) with a small percentage of training data.}
    \label{fig:sample efficiency}
\end{figure*}

\subsection{Setup}
\textbf{Nuscenes dataset.} \cite{nuscenes2019} is a dataset for autonomous driving based in Boston and Singapore. It contains 850 scenes each 20s long (sampled at 2hz, therefore a max of 40 steps), containing 23 object classes and HD semantic maps with 11 annotated layers. We chose this particular dataset for the rich semantics it provides which is well suited for rule definitions. We will use 650 scenes for training and 200 scenes for validation. During close-loop evaluation, all agents are rolled out synchronously and the ego agent's motion is controlled by our neural controller.

\textbf{Methods of evaluation.} We evaluate our method and comparison cases with the following metrics. \textit{Percentage of close encounters} measures the average of times the ego vehicles comes to the vicinity of ado vehicles in a scene (safety measure). \textit{Acceleration} is the maximum magnitude acceleration during a scene (comfort measure). \textit{Similarity to human driving} is the L2-norm between the planner and human trajectories (driving style measure). \textit{Goal distance} is the ego's final distance to the goal. All metrics are the lower the better. During evaluation, we control the ego vehicle with our learned planner, the ado vehicles move according to the trajectories recorded in the dataset with synchronized time. All results are averaged over the validation set.

It is worth mentioning that, the four evaluation metrics that we use are more trade-offs than objectives that can be optimized concurrently. For example, a controller that achieves a low goal distance over the evaluation scenes (within the fixed time horizon) is bound to drive fast and hence obtain a higher percentage close encounter and acceleration. Similarly, a controller that achieves high human driving similarity may obtain moderate scores on the other metrics. In our case, our objective function is the MSE loss with respective to the human ego vehicle's controls, therefore we put high emphasis on human driving similarity. In general planning and control scenarios, it ultimately depends on the users' preferences (which to a certain level can be controlled by tuning $\alpha, \beta$).

\textbf{Comparison cases.} Five planner variants are used for comparison. \textit{Ours} refers to the proposed method; \textit{Human} refers the human driver in the dataset; \textit{CNN} refers to a planner that maps BEV directly to controls (Implemented similarly to \cite{Farag2018BehaviorCF}); and \textit{CNN-LSTM} refers the previous planner with an added LSTM component to keep track of history. \textit{RvS-G} refers to the goal conditioned offline RL via supervised learning proposed in \cite{rvs}. For all planners, the same CNN backbone is used to extract features from the rasterized BEV image (similar to \cite{Cui2019MultimodalTP}). For all cases other than \textit{Human}, we use the same CNN backbone to process the input BEV image. In the table, FC denotes fully connected layer, F denotes number of filters, K denotes kernel size, S denotes stride, U denotes number of units in the fully connected layer. For \textit{CNN}, we concatenate the CNN features with the goal pose, which are passed through 2 FC layers that output speed and steering. For \textit{CNN-LSTM}, the flattened features of the CNN backbone with concatenated goal pose are passed through 1 FC layer. The output of the FC layer is passed to an LSTM (with 64 dimensional hidden state) as input, which in turn provides speed and steering. This is a modified architecture of \cite{farag2018behavior} that is used for behavior cloning of self-driving policies. RvS-G refers to the goal conditioned offline RL via supervised learning proposed in \cite{rvs} which has been shown to outperform a number of behavior cloning and RL methods. In our implementation, we pass the goal and the current BEV image into the same CNN feature extractor, the concatenated feature vector is passed to the LSTM for control generation. In addition to providing the current BEV image, we also provide the goal BEV image as input.


\subsection{Results and discussion}

\begin{figure*}[htb]
    \centering
    \includegraphics[width=0.8\linewidth]{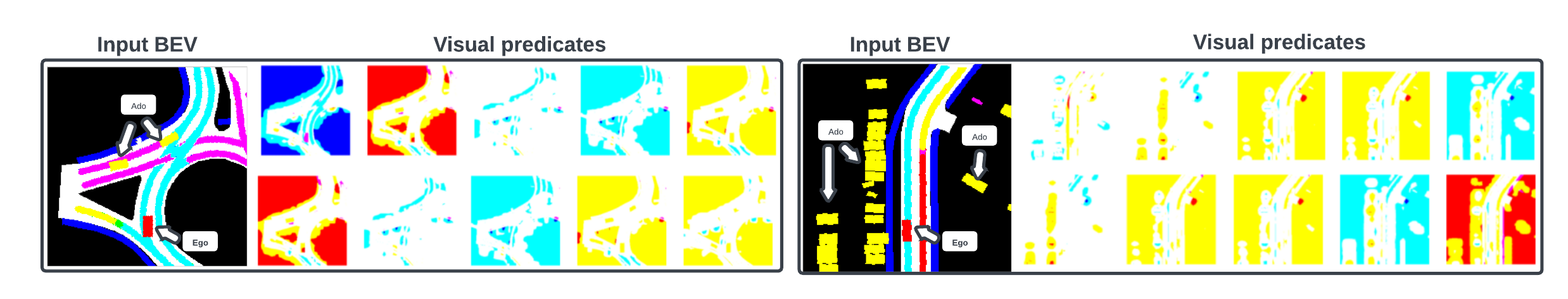}
    \caption{\textbf{Visual predicate feature maps.}  visualizations of the feature maps (overlayed on the input BEV image) from the visual predicate extractor. Color indicates the level of attention and is ranks from high to low as: $\textrm{yellow} \rightarrow \textrm{red} \rightarrow \textrm{cyan} \rightarrow \textrm{magenta}$}
    \label{fig:q vp map}
\end{figure*}

 \begin{figure}[!htb]
    \centering
    \includegraphics[width=0.8\linewidth]{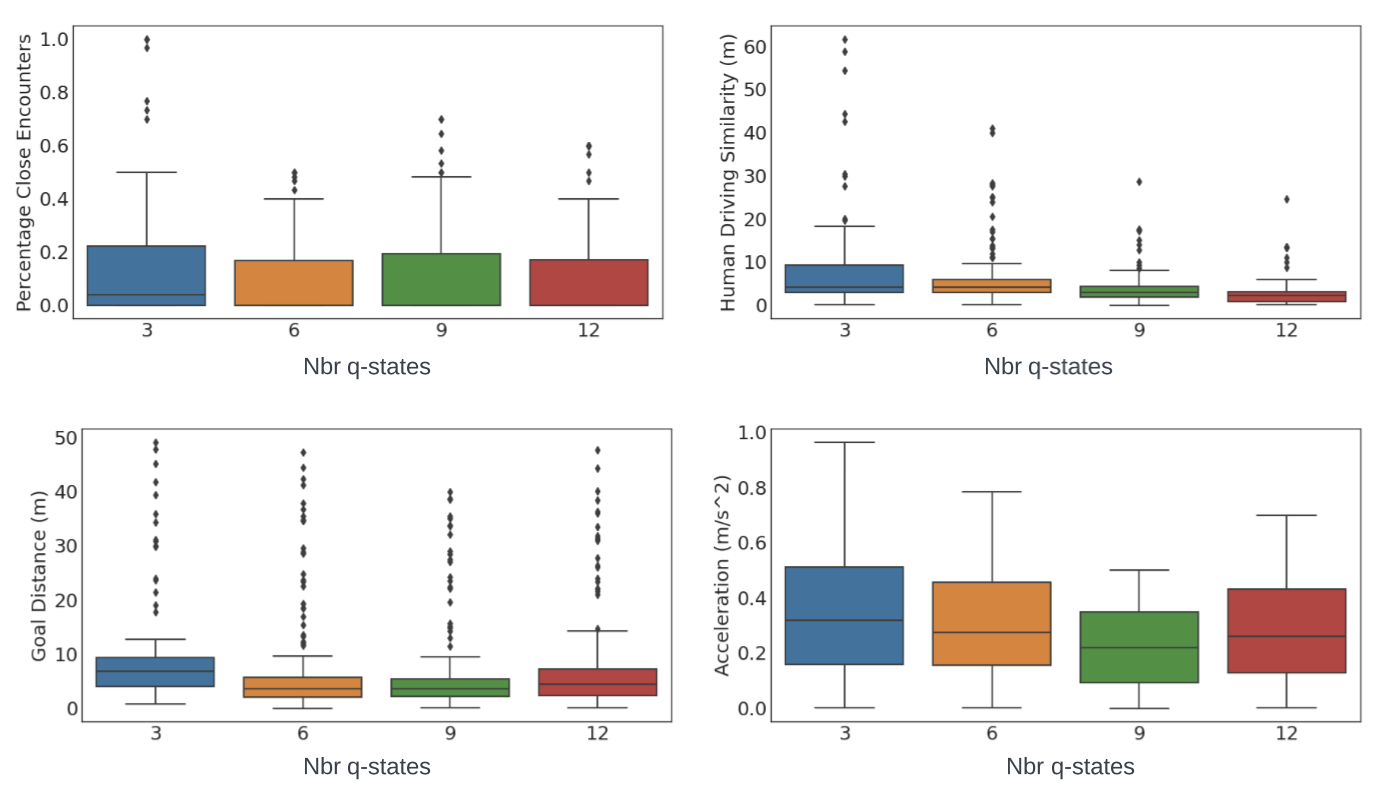}
    \caption{\textbf{Study on varying the number of q-states}. With increasing number of q-states, human driving similarity considerably improves. Beyond a minimal number of q-states, performance of vAGN can improve but tuning is required to obtain the desired balance between performance and explainability.}\label{fig:q-node study}
\end{figure}

\textbf{vAGN learns an explainable behavior planner that attends to semantically meaningful regions on the BEV.} Figure \ref{fig:Execution traces} illustrates 3 time-steps during navigation of a roundabout. Within each sub-figure we also show the current vAGN state distribution (darker the color means higher probability) and the saliency map indicating where each q-state is attending to. Color indicates the level of attention and is ranks from high to low as: $\textrm{yellow} \rightarrow \textrm{red} \rightarrow \textrm{cyan} \rightarrow \textrm{magenta}$. To the right of each q-distribution plot we also show the semantically colored BEV image sent as input to the network. In Figure \ref{fig:Execution traces}(a)(b), the ego vehicle is driving in an area of sparse traffic, the 2 vAGN nodes with highest probabilities are $q_2$ and $q_3$ both of which focuses on the undrivable areas which helps to prevent collision with those areas. As the ego comes to an intersection with ado vehicles in front (Figure \ref{fig:Execution traces}(c)), some probabilities shift from $q_2$ to $q_3$ where $q_3$ puts more attention of the ado vehicles. Finally, as the ego vehicle navigates out of the aroundabout (Figure \ref{fig:Execution traces}(d)), the q-states goes back to attending the undrivable areas. Please see the video attachment for full execution runs. Note that The meaning of each q-state depends on the training and the q-states may not always possess clear semantic meanings in human terms (i.e. turning, accelerating, etc). The execution trace in Figure \ref{fig:Execution traces} shows that q2 corresponds to avoiding collision with the road boundary (more weight on q3 when the ego vehicle comes close to the road boundary). Whereas q3 corresponds to avoiding vehicle-to-vehicle collision (from Figure 3-b to 3-c, q3’s weighting increased as the ego vehicle comes close to the ados vehicles in the roundabout.)

\textbf{vAGN learns to modulate DMP parameters to exhibit safe behaviors.} Recall in Section \ref{sec:vAGN_DMP} we described that the output of vAGN is used to control the DMP parameters which in turn controls the ``aggressiveness" of goal reaching. In Figure \ref{fig:Explainability traces} we show an example of how this is achieved. Figure \ref{fig:Explainability traces} (a) shows the evolution of the DMP parameters within a scene execution. It is noticeable that there is an abrupt change in the parameters starting from step 15. Figure \ref{fig:Explainability traces}(d) shows the BEV input to the network, during steps 15, 16 and 17 the ego vehicle is trying to make a left turn but comes very close to the boundry of the road, therefore vAGN decides to focus on the road boundries (shown by highest weighted nodes $q_0, q_1, q_4$ and their corresponding saliency maps), which controls the change in the DMP parameters. Figure \ref{fig:Explainability traces}(c) shows how the shift in q-distribution effects the resultant speed and steering (controls). From step 15 onwards, this shift has resulted in a mild speed decrease and more aggressive steering relieve, both of which steers the ego vehicle away from the road boundary. Refer to the video supplementary for a complete execution of this scene (along with others).

\textbf{vAGN strikes a balance between safety and comfort, and  exhibits near human driving behavior.} Table \ref{tab:1} shows the close-loop rollout performance of all comparison cases over the validation set. We evaluated each model with 3 random seeds and Table \ref{tab:1} reports the mean performance with standard deviation. Our method is able to achieve the best goal reaching performance and similarity to human driving (ADE). Because each scene rollout is fixed maximum time-step (20 seconds sampled at 2hz or 40 steps defined by the dataset), driving conservatively (high safety and comfort scores) is a trade-off to goal achievement. We can see that \textit{CNN} and \textit{RvS-G} achieves the best safety and comfort scores respectively but in turn performs less than ideal in other metrics. In comparison, our method exhibits safety and comfort level similar to that of the human driver in the dataset.

\begin{table}[]
\centering
\caption{Performance Comparison}
\label{tab:1}
\begin{tabular}{|c|cl|cl|}
\hline
\multirow{2}{*}{\textbf{Model}} & \multicolumn{2}{c|}{\textbf{\%Close Encounters}} & \multicolumn{2}{c|}{\textbf{Acceleration}} \\ \cline{2-5} 
 & \multicolumn{2}{c|}{\textbf{mean $\pm$ std}} & \multicolumn{2}{c|}{\textbf{mean $\pm$ std}} \\ \hline
\textbf{Human} & \multicolumn{2}{c|}{19.0\%} & \multicolumn{2}{c|}{0.30} \\ \hline
\textbf{CNN} & \multicolumn{2}{c|}{\textbf{11.0\% $\pm$ 2.0\%}} & \multicolumn{2}{c|}{0.92 $\pm$ 0.32} \\ \hline
\textbf{CNN-LSTM} & \multicolumn{2}{c|}{12.2\% $\pm$ 3.2\%} & \multicolumn{2}{c|}{1.20 $\pm$ 0.25} \\ \hline
\textbf{RvS-G} & \multicolumn{2}{c|}{26.8\% $\pm$ 5.0\%} & \multicolumn{2}{c|}{\textbf{0.22 $\pm$ 0.05}} \\ \hline
\textbf{Ours} & \multicolumn{2}{c|}{16.2\% $\pm$ 1.3\%} & \multicolumn{2}{c|}{0.43 $\pm$ 0.10} \\ \hline
\multirow{2}{*}{\textbf{Model}} & \multicolumn{2}{c|}{\textbf{Human Driving Sim.}} & \multicolumn{2}{c|}{\textbf{Goal Distance}} \\ \cline{2-5} 
 & \multicolumn{2}{c|}{\textbf{mean $\pm$ std}} & \multicolumn{2}{c|}{\textbf{mean $\pm$ std}} \\ \hline
\textbf{Human} & \multicolumn{2}{c|}{n/a} & \multicolumn{2}{c|}{n/a} \\ \hline
\textbf{CNN} & \multicolumn{2}{c|}{25.30 $\pm$ 8.20} & \multicolumn{2}{c|}{14.20 $\pm$ 2.80} \\ \hline
\textbf{CNN-LSTM} & \multicolumn{2}{c|}{19.23 $\pm$ 5.30} & \multicolumn{2}{c|}{19.80 $\pm$ 3.10} \\ \hline
\textbf{RvS-G} & \multicolumn{2}{c|}{14.25 $\pm$ 2.80} & \multicolumn{2}{c|}{32.91 $\pm$ 6.30} \\ \hline
\textbf{Ours} & \multicolumn{2}{c|}{\textbf{5.87 $\pm$ 1.40}} & \multicolumn{2}{c|}{\textbf{8.20 $\pm$ 1.50}} \\ \hline
\end{tabular}
\end{table}

\textbf{The multi-abstractive neural controller achieves high sample efficiency.} Because of the DMP, our neural controller already has a level of lane following capabilities built-in (even before any training). Given this structure, we can expect our controller to exhibit improved sample efficiency. This is shown in Figure \ref{fig:sample efficiency}(a) where we train on 4 different levels sub-training set (evaluation is done on the full validation set). The results show that our model is able to achieve relatively high human driving similarity (low average displacement error) even when trained with a small training set. Because of its structure, the size of our model is also significantly smaller (more than 30\% less parameters) than comparison models as shown in Figure \ref{fig:sample efficiency}(b). This is important for planning and control modules as smaller models promote higher inference frequencies at runtime.

\begin{figure}[!htb]
\centering
\includegraphics[width=0.8\linewidth]{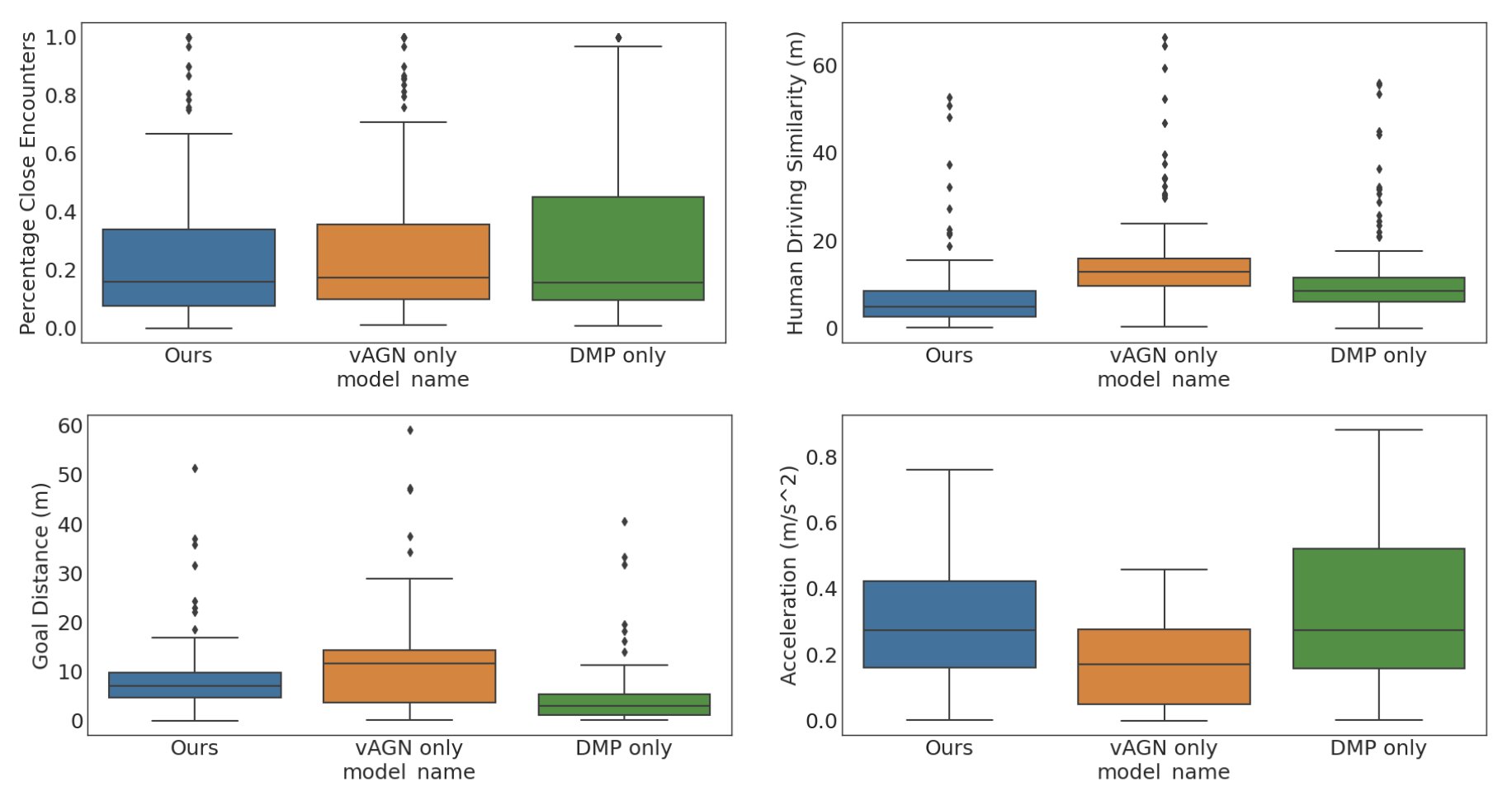}
\caption{\textbf{Self-ablation study.} vAGN helps in learning coarse and interactive maneuvers whereas DMP is responsible for accurate and sample efficient navigation (e.g. reference path following). The combination of the two addresses different components of driving}
\label{fig:self abalation}
\end{figure}

\textbf{vAGN facilitates the CNN feature extractor to learn semantically meaningful visual predicates}. Figure \ref{fig:q vp map} shows visualizations of the feature maps (overlayed on the input BEV image) from the visual predicate extractor. Color indicates the level of attention and is ranks from high to low as: $\textrm{yellow} \rightarrow \textrm{red} \rightarrow \textrm{cyan} \rightarrow \textrm{magenta}$. In both examples, we can see that the visual predicates attend to semantically meaningful components of the road (lanes, road boundaries, ado vehicles, etc). For the current design of the visual predicate extractor, we have found empirically that the objects each visual predicate is attending to evolves over time (during execution) and there are at times duplications (as shown in both figures). Looking into this behavior will be future work. Overall, these predicates and their corresponding feature vector provides the basis for vAGN to perform appropriate transitions which in turn controls the DMP to output interactive behaviors.

\textbf{Altering the number of q-states trade off between explainability and performance}. We treat the number of q-states as a hyperparameter that requires tuning. The trade-off is between explainability and performance (vAGN with less q-states are more explainable, may learn more semantically meaningful driving modes at each node, but with reduced overall performance). To study the effect of vAGN size on the neural controller, we conducted a study where the number of q-states are varied from 3 to 12 at increments of 3 and the results are presented in Figure \ref{fig:q-node study}. The trend that stood out is that with increasing number of q-states, human driving similarity improves significantly. This is because as number of weights (hence the representative power) of vAGN improves as the number q-states increase. The performance of other metrics also improves with the number of q-states (especially when $\#q>3$). However, the ego vehicle starts to drive more aggressively at $\#q=12$ as shown by the jump in acceleration. In summary, beyond a minimal number of q-states, performance of vAGN can improve but tuning is required to obtain the desired balance between performance and explainability.

\textbf{Within the multi-abstractive neural controller, vAGN takes care of safety and interactive behaviors and DMP is responsible for driving along lanes}. In Figure \ref{fig:self abalation}, we study the influence of the two main components - vAGN and DMP on our hierarchical neural controller. As a reminder, our controller follows a $CNN \rightarrow vAGN \rightarrow DMP$ structure. In the figure, \textit{Ours} refers to the architecture with both components. \textit{vAGN only} refers to $CNN \rightarrow vAGN \rightarrow FC$. \textit{DMP only} refers to $CNN \rightarrow DMP$. The results for \textit{DMP only} shows that DMP contributes most to goal reaching and human driving similarity, which is reasonable because DMP  takes the main responsibility for lane following. However, with only DMP, the ego vehicle drives aggressively (shown in acceleration and percentage close encounter to nearby cars). Meanwhile, using vAGN improves the safety aspect of driving (lower acceleration and close encounter) because it learns to attend to road boundaries and ado vehicles and issues commands to avoid them. But vAGN alone performs less ideally in reaching the goal and human driving similarity is also worse. In a nutshell, vAGN helps in learning coarse and interactive maneuvers whereas DMP is useful for accurate navigation (e.g. reference path following). The combination of the two addresses different components of driving.



\section{Conclusion}
\label{sec:conclusion}

In this work, we introduced the multi-abstractive neural controller which is a differentiable representation of a simplified plan/control stack. Within which we introduced the visual automaton generative network that acts as a behavior with learnable structure. We show that just by using supervised learning this policy representation is able to achieve high sample efficiency and a well balanced performance in terms of safety, optimality and comfort. Because of its structure, the decision making process of vAGN is highly interpretable. We show that it learns to attend to semantically meaningful regions of the input image while making transitions among (learned) modes of operations.

\bibliographystyle{IEEEtran}
\bibliography{references.bib}

\begin{thebibliography}{10}
\providecommand{\url}[1]{#1}
\csname url@samestyle\endcsname
\providecommand{\newblock}{\relax}
\providecommand{\bibinfo}[2]{#2}
\providecommand{\BIBentrySTDinterwordspacing}{\spaceskip=0pt\relax}
\providecommand{\BIBentryALTinterwordstretchfactor}{4}
\providecommand{\BIBentryALTinterwordspacing}{\spaceskip=\fontdimen2\font plus
\BIBentryALTinterwordstretchfactor\fontdimen3\font minus
  \fontdimen4\font\relax}
\providecommand{\BIBforeignlanguage}[2]{{%
\expandafter\ifx\csname l@#1\endcsname\relax
\typeout{** WARNING: IEEEtran.bst: No hyphenation pattern has been}%
\typeout{** loaded for the language `#1'. Using the pattern for}%
\typeout{** the default language instead.}%
\else
\language=\csname l@#1\endcsname
\fi
#2}}
\providecommand{\BIBdecl}{\relax}
\BIBdecl

\bibitem{Zeng2019EndToEndIN}
W.~Zeng, W.~Luo, S.~Suo, A.~Sadat, B.~Yang, S.~Casas, and R.~Urtasun,
  ``End-to-end interpretable neural motion planner,'' \emph{CVPR}, pp.
  8652--8661, 2019.

\bibitem{Salzmann2022NeuralMPCDL}
T.~Salzmann, E.~Kaufmann, M.~Pavone, D.~Scaramuzza, and M.~Ryll, ``Neural-mpc:
  Deep learning model predictive control for quadrotors and agile robotic
  platforms,'' \emph{ArXiv}, vol. abs/2203.07747, 2022.

\bibitem{learnmppi}
J.~Sacks and B.~Boots, ``Learning to optimize in model predictive control,''
  \emph{ICRA}, 2022.

\bibitem{Ichnowski2021AcceleratingQO}
J.~Ichnowski, P.~Jain, B.~Stellato, G.~Banjac, M.~Luo, F.~Borrelli, J.~E.
  Gonzalez, I.~Stoica, and K.~Goldberg, ``Accelerating quadratic optimization
  with reinforcement learning,'' in \emph{NeurIPS}, 2021.

\bibitem{Dawson2022SafeCW}
C.~Dawson, S.~Gao, and C.~Fan, ``Safe control with learned certificates: A
  survey of neural lyapunov, barrier, and contraction methods,'' \emph{ArXiv},
  vol. abs/2202.11762, 2022.

\bibitem{Ijspeert2002MovementIW}
A.~J. Ijspeert, J.~Nakanishi, and S.~Schaal, ``Movement imitation with
  nonlinear dynamical systems in humanoid robots,'' \emph{ICRA}, vol.~2, pp.
  1398--1403 vol.2, 2002.

\bibitem{Meng2019NeuralAN}
X.~Meng, N.~D. Ratliff, Y.~Xiang, and D.~Fox, ``Neural autonomous navigation
  with riemannian motion policy,'' \emph{ICRA}, pp. 8860--8866, 2019.

\bibitem{Kabzan2019LearningBasedMP}
J.~Kabzan, L.~Hewing, A.~Liniger, and M.~N. Zeilinger, ``Learning-based model
  predictive control for autonomous racing,'' \emph{IEEE Robotics and
  Automation Letters}, vol.~4, pp. 3363--3370, 2019.

\bibitem{prm-rl}
A.~Faust, O.~Ramirez, M.~Fiser, K.~Oslund, A.~G. Francis, J.~Davidson, and
  L.~Tapia, ``Prm-rl: Long-range robotic navigation tasks by combining
  reinforcement learning and sampling-based planning,'' \emph{2018 IEEE
  International Conference on Robotics and Automation (ICRA)}, pp. 5113--5120,
  2018.

\bibitem{Fox2019Multi}
R.~Fox, R.~Berenstein, I.~Stoica, and K.~Goldberg, ``Multi-task hierarchical
  imitation learning for home automation,'' in \emph{CASE}, 2019.

\bibitem{class_activation_map}
B.~Zhou, A.~Khosla, {\`A}.~Lapedriza, A.~Oliva, and A.~Torralba, ``Learning
  deep features for discriminative localization,'' \emph{2016 IEEE Conference
  on Computer Vision and Pattern Recognition (CVPR)}, pp. 2921--2929, 2015.

\bibitem{Bojarski2016VisualBackPropVC}
M.~Bojarski, A.~Choromańska, K.~Choromanski, B.~Firner, L.~D. Jackel,
  U.~Muller, and K.~Zieba, ``Visualbackprop: visualizing cnns for autonomous
  driving,'' \emph{ArXiv}, vol. abs/1611.05418, 2016.

\bibitem{Lechner2020NeuralCP}
M.~Lechner, R.~M. Hasani, A.~Amini, T.~A. Henzinger, D.~Rus, and R.~Grosu,
  ``Neural circuit policies enabling auditable autonomy,'' \emph{Nature Machine
  Intelligence}, vol.~2, pp. 642--652, 2020.

\bibitem{Hudson2019LearningBA}
D.~A. Hudson and C.~D. Manning, ``Learning by abstraction: The neural state
  machine,'' in \emph{NeurIPS}, 2019.

\bibitem{Kochiev2021NeuralSM}
L.~Kochiev, ``Neural state machine for {2D} and {3D} visual question
  answering,'' 2021.

\bibitem{Hannun2020DifferentiableWF}
A.~Y. Hannun, V.~Pratap, J.~Kahn, and W.-N. Hsu, ``Differentiable weighted
  finite-state transducers,'' \emph{ArXiv}, vol. abs/2010.01003, 2020.

\bibitem{Ke2022LearningTI}
N.~R. Ke, S.~Chiappa, J.~X. Wang, J.~Bornschein, T.~Weber, A.~Goyal,
  M.~Botvinic, M.~C. Mozer, and D.~J. Rezende, ``Learning to induce causal
  structure,'' \emph{ArXiv}, vol. abs/2204.04875, 2022.

\bibitem{Starke2019NeuralSM}
S.~Starke, H.~Zhang, T.~Komura, and J.~Saito, ``Neural state machine for
  character-scene interactions,'' \emph{ACM Transactions on Graphics (TOG)},
  vol.~38, pp. 1 -- 14, 2019.

\bibitem{Poli2021NeuralHA}
M.~Poli, S.~Massaroli, L.~Scimeca, S.~J. Oh, S.~Chun, A.~Yamashita, H.~Asama,
  J.~Park, and A.~Garg, ``Neural hybrid automata: Learning dynamics with
  multiple modes and stochastic transitions,'' in \emph{NeurIPS}, 2021.

\bibitem{Lipton2016TheMO}
Z.~C. Lipton, ``The mythos of model interpretability,'' \emph{Queue}, vol.~16,
  pp. 31 -- 57, 2016.

\bibitem{Narayanan2018HowDH}
M.~Narayanan, E.~Chen, J.~He, B.~Kim, S.~J. Gershman, and F.~Doshi-Velez, ``How
  do humans understand explanations from machine learning systems? an
  evaluation of the human-interpretability of explanation,'' \emph{ArXiv}, vol.
  abs/1902.00006, 2018.

\bibitem{Ijspeert2013DynamicalMP}
A.~J. Ijspeert, J.~Nakanishi, H.~Hoffmann, P.~Pastor, and S.~Schaal,
  ``Dynamical movement primitives: Learning attractor models for motor
  behaviors,'' \emph{Neural Computation}, vol.~25, pp. 328--373, 2013.

\bibitem{AbuDakka2015AdaptationOM}
F.~J. Abu-Dakka, B.~Nemec, J.~A. J{\o}rgensen, T.~R. Savarimuthu,
  N.~Kr{\"u}ger, and A.~Ude, ``Adaptation of manipulation skills in physical
  contact with the environment to reference force profiles,'' \emph{Autonomous
  Robots}, vol.~39, pp. 199--217, 2015.

\bibitem{Lin2014NetworkIN}
M.~Lin, Q.~Chen, and S.~Yan, ``Network in network,'' \emph{CoRR}, vol.
  abs/1312.4400, 2014.

\bibitem{Zhou2016LearningDF}
B.~Zhou, A.~Khosla, {\`A}.~Lapedriza, A.~Oliva, and A.~Torralba, ``Learning
  deep features for discriminative localization,'' \emph{2016 IEEE Conference
  on Computer Vision and Pattern Recognition (CVPR)}, pp. 2921--2929, 2016.

\bibitem{Koutras2019ACF}
L.~Koutras and Z.~Doulgeri, ``A correct formulation for the orientation dynamic
  movement primitives for robot control in the cartesian space,'' in
  \emph{CoRL}, 2019.

\bibitem{nuscenes2019}
H.~Caesar, V.~Bankiti, A.~H. Lang, S.~Vora, V.~E. Liong, Q.~Xu, A.~Krishnan,
  Y.~Pan, G.~Baldan, and O.~Beijbom, ``nuscenes: A multimodal dataset for
  autonomous driving,'' \emph{arXiv preprint arXiv:1903.11027}, 2019.

\bibitem{Farag2018BehaviorCF}
W.~Farag and Z.~Saleh, ``Behavior cloning for autonomous driving using
  convolutional neural networks,'' \emph{2018 International Conference on
  Innovation and Intelligence for Informatics, Computing, and Technologies
  (3ICT)}, pp. 1--7, 2018.

\bibitem{rvs}
S.~Emmons, B.~Eysenbach, I.~Kostrikov, and S.~Levine, ``Rvs: What is essential
  for offline rl via supervised learning?'' \emph{arXiv preprint
  arXiv:2112.10751}, 2021.

\bibitem{Cui2019MultimodalTP}
H.~Cui, V.~Radosavljevic, F.-C. Chou, T.-H. Lin, T.~Nguyen, T.-K. Huang,
  J.~Schneider, and N.~Djuric, ``Multimodal trajectory predictions for
  autonomous driving using deep convolutional networks,'' \emph{ICRA}, pp.
  2090--2096, 2019.

\bibitem{farag2018behavior}
W.~Farag and Z.~Saleh, ``Behavior cloning for autonomous driving using
  convolutional neural networks,'' in \emph{2018 International Conference on
  Innovation and Intelligence for Informatics, Computing, and Technologies
  (3ICT)}.\hskip 1em plus 0.5em minus 0.4em\relax IEEE, 2018, pp. 1--7.

\end{thebibliography}

\end{document}